\pdfoutput=1
\documentclass[11pt]{article}

\usepackage[final]{acl}

\usepackage{times}
\usepackage{latexsym}
\usepackage{makecell}
\usepackage[T1]{fontenc}
\usepackage[utf8]{inputenc}

\usepackage{microtype}
\usepackage{multirow}
\usepackage{caption}
\usepackage{inconsolata}

\usepackage{graphicx}

\title{Focused ReAct: Improving ReAct through Reiterate and Early Stop}

\author{
  Shuoqiu Li \\
  Carnegie Mellon University \\
  \texttt{shuoqiul@alumni.cmu.edu} 
  \And
  Han Xu \\
  University of Illinois at Urbana-Champaign \\
  \texttt{hanxu8@illinois.edu}
  \AND
  Haipeng Chen \\
  William \& Mary \\
  \texttt{hchen23@wm.edu}
}

\begin{document}
\maketitle
\begin{abstract}
Large language models (LLMs) have significantly improved their reasoning and decision-making capabilities, as seen in methods like ReAct. However, despite its effectiveness in tackling complex tasks, ReAct faces two main challenges: losing focus on the original question and becoming stuck in action loops. To address these issues, we introduce \textbf{Focused ReAct}, an enhanced version of the ReAct paradigm that incorporates reiteration and early stop mechanisms. These improvements help the model stay focused on the original query and avoid repetitive behaviors. Experimental results show accuracy gains of 18\% to 530\% and a runtime reduction of up to 34\% compared to the original ReAct method.
\end{abstract}

\section{Introduction}

Recent advancements in large language models (LLMs) have enabled more sophisticated techniques for reasoning and decision-making. One such technique, the ReAct framework (Reason+Act), has gained popularity for its dual approach of alternating between reasoning and action~\cite{yao2023react}. This combination allows ReAct to excel in handling complex tasks by better adapting to dynamic environments~\cite{wang2024survey}.

\begin{figure}[t]
  \includegraphics[width=\columnwidth]{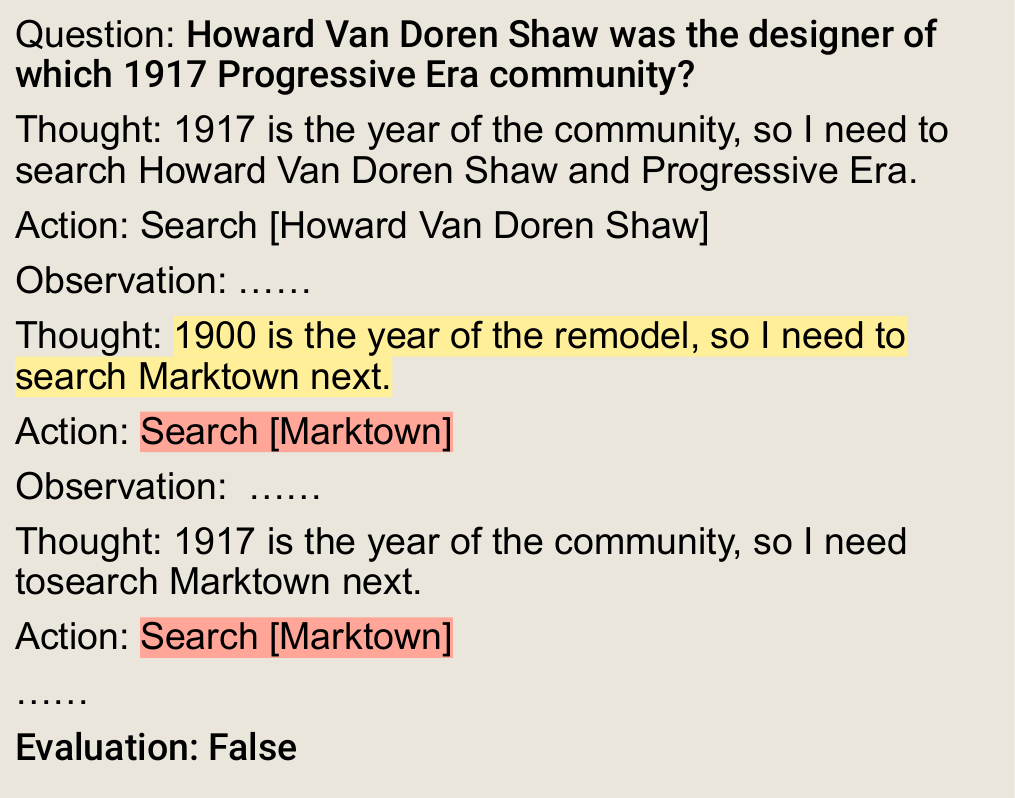}
  \caption{The yellow highlighted region illustrates where the model loses focus on the original question, while the red highlighted region depicts repeated actions that result in a failed response.}
  \label{fig:issue}
\end{figure}

\label{sec:intro}
Despite its strengths in general question-answering (QA), ReAct sometimes falls short in delivering accurate results, as demonstrated in Figure~\ref{fig:issue}. When confronted with lengthy or intricate questions, the model—paired with the ReAct framework—may lose track of the original question, leading to incorrect responses. Additionally, the ReAct method can get trapped in repetitive action loops, continuously performing the same action without progressing toward a solution. To address these challenges, we introduce \textbf{Focused ReAct}, an enhanced version of the original ReAct method specifically designed for QA tasks.

\section{Methodology}

This section outlines the two core features of the \textbf{Focused ReAct} method: \textbf{reiterate} and \textbf{early stop}.

\subsection{Reiterate to Solve Context Loss}

As indicated in the introduction, The first challenge stems from the extended reasoning process, where the relevance of the original question diminishes as additional reasoning and actions are appended. To resolve this, the concept of \textbf{reiterate} is introduced. In this approach, the original question is restated at the beginning of each reasoning step in the ReAct cycle. This technique is illustrated in the green-highlighted region of Figure~\ref{fig:issue2}.

By reiterating the original question at each step, the model continually emphasizes the user’s query, preventing it from being overshadowed by the increasingly long context that ReAct tends to create. This simple yet effective strategy mitigates the context dilution problem illustrated in Figure~\ref{fig:issue}, ensuring that the output remains aligned with the user’s request, even in complex or multi-step tasks.

\subsection{Early Stop to Prevent Action Repetition}

The second challenge, as outlined in the introduction, occurs when the model gets caught in repetitive loops, generating the same response without progressing toward the correct answer. To tackle this, we propose an \textbf{early stop} mechanism. It assumes that by the time a duplicate action occurs, sufficient information has been gathered. 

\begin{figure}[t]
  \includegraphics[width=\columnwidth]{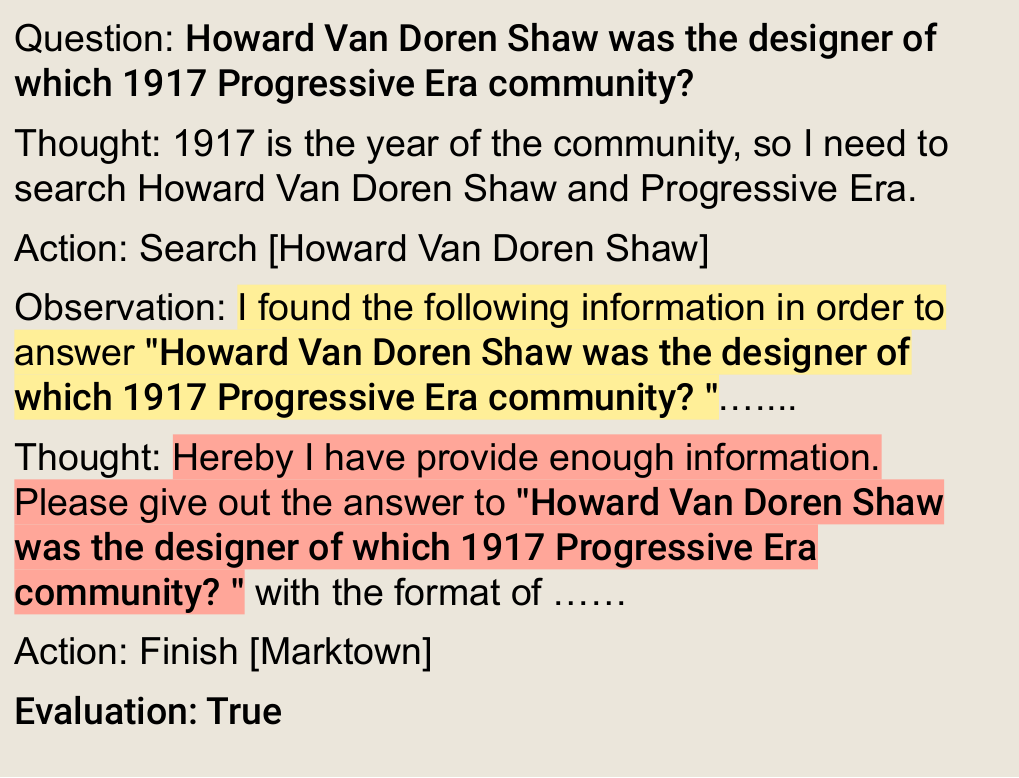}
  \caption{The QA process by Focused ReAct for the same question, which applies reiteration (highlighted in yellow) and early stop (highlighted in red) to resolve the context loss and the repeated action issue.}
  \label{fig:issue2}
\end{figure}

When the program detects repeated actions, it triggers a termination request - highlighted in red in Figure~\ref{fig:issue2} - instructing the model to generate a final answer based on the existing information. This approach prevents unnecessary repetition and helps the QA process arrive at an accurate response more efficiently.

\section{Experimentation}
We evaluate Focused ReAct against the ReAct baseline using the Gemma 2 2B~\cite{team2024gemma}, Phi-3.5-mini 3.8B~\cite{abdin2024phi} and Llama 3.1 8B~\cite{dubey2024llama} models. The implementation uses the PyTorch and Transformers libraries\footnote{Our
code implementation and experiments are available at \href{https://github.com/wmd3i/Focused-ReAct}{https://github.com/wmd3i/Focused-ReAct}.}, with experiments conducted on a single NVIDIA L4 GPU with 24GB of memory. The dataset consists of 150 QA tasks, randomly selected from HotPotQA~\cite{yang2018hotpotqa}. We measure accuracy as the ratio of correctly answered tasks to the total number of tasks, while runtime is recorded for the completion of each task.

\begin{table}[ht]
    \centering
    \caption{Accuracy Comparison of ReAct vs. Focused ReAct}
    \fontsize{8}{10}\selectfont
    \begin{tabular}{|c|c|c|c|}
        \hline
        Model & ReAct & \makecell{Focused \\ ReAct} & abs./rel. diff \\
        \hline
        Gemma 2 2B & 2.0\% & \textbf{12.6\%} & +\textbf{10.6} / \textbf{530\%}\\
        Phi-3.5-mini 3.8B & 22.0\% & \textbf{26.0\%} & +4.0 / 18\%\\
        Llama 3.1 8B & 14.0\% & \textbf{23.3\%} & +9.3 / 66\%\\
        \hline
    \end{tabular}
    \label{tab:accuracy}
\end{table}

Table~\ref{tab:accuracy} presents the accuracy comparison between the vanilla ReAct and Focused ReAct across the Gemma 2, Phi-3.5, and Llama 3.1 models. Focused ReAct demonstrates an 18\%-530\% improvement in accuracy.

\begin{table}[h]
    \centering
    \caption{Runtime Comparison (Average and Std) for ReAct vs. Focused ReAct}
    \fontsize{8}{10}\selectfont
    \begin{tabular}{|c|c|c|c|}
        \hline
        Model & ReAct & \makecell{Focused \\ ReAct} & \makecell{abs./rel. \\ diff} \\
        \hline
        Gemma 2 2B  & 11.68$\pm$2.66s  & \textbf{7.68}$\pm$\textbf{2.41s} & -\textbf{4.0} / \textbf{34}\% \\
        \makecell{Phi-3.5\\-mini 3.8B}  & 23.23$\pm$8.42s & \textbf{22.50}$\pm$\textbf{11.19}s & -0.73 / 3\% \\
        Llama 3.1 8B  & 24.10$\pm$23.48s & \textbf{23.12}$\pm$\textbf{25.35}s & -0.98 / 4\% \\
        \hline
    \end{tabular}
    \label{tab:runtime}
\end{table}

Table~\ref{tab:runtime} summarizes the average runtime and standard deviation (std) for both the original ReAct and Focused ReAct methods. Models with fewer parameters show a 34\% reduction in runtime, while models with larger parameter sizes exhibit no significant decrease. This discrepancy may be attributed to the fact that smaller models, with weaker reasoning capabilities, benefit more from Focused ReAct optimizations. In contrast, larger models are more robust at maintaining context and performing deeper reasoning, which may reduce the relative impact of Focused ReAct’s efficiency gains. As a result, the runtime benefits are less pronounced compared to smaller models.

\section{Conclusion}

This paper identifies two common issues with the ReAct method in QA: losing focus on the original question during extended reasoning and becoming stuck in repetitive action loops. To overcome these problems, we propose \textbf{Focused ReAct}, which incorporates \textbf{reiteration} and \textbf{early stop} to improve upon the ReAct framework. Compared to the original ReAct method, the new approach achieves accuracy improvements between 18\% and 530\%, along with a reduction in runtime of up to 34\%.

For future work, we plan to extend Focused ReAct to a broader range of tasks and scenarios, evaluate its generalizability and robustness, and explore techniques to further accelerate its performance~\cite{xu2024can}.

\newpage
\bibliography{custom}

\end{document}